\documentclass[10pt,twocolumn,letterpaper]{article}

\usepackage{symbols}
\usepackage[pagenumbers]{iccv} 

\definecolor{iccvblue}{rgb}{0.21,0.49,0.74}
\usepackage{colortbl}
\definecolor{lightgray}{gray}{0.9}
\usepackage[pagebackref,breaklinks,colorlinks,allcolors=iccvblue]{hyperref}

\def\paperID{8973} 
\def\confName{ICCV}
\def\confYear{2025}

\title{VALLR: Visual ASR Language Model for Lip Reading}

\author{
Marshall Thomas \quad Edward Fish \quad Richard Bowden \\
University of Surrey \\
\texttt{mt00893@surrey.ac.uk} \quad 
\texttt{ef0036@surrey.ac.uk} \quad 
\texttt{r.bowden@surrey.ac.uk}
}

\begin{document}
\maketitle
\begin{abstract}
Lip Reading, or Visual Automatic Speech Recognition (V-ASR), is a complex task requiring the interpretation of spoken language exclusively from visual cues, primarily lip movements and facial expressions. This task is especially challenging due to the absence of auditory information and the inherent ambiguity when visually distinguishing phonemes that have overlapping visemes, where different phonemes appear identical on the lips. Current methods typically attempt to predict words or characters directly from these visual cues, but this approach frequently encounters high error rates due to coarticulation effects and viseme ambiguity. We propose a novel two-stage, phoneme-centric framework for Visual Automatic Speech Recognition (V-ASR) that addresses these longstanding challenges. First, our model predicts a compact sequence of phonemes from visual inputs using a Video Transformer with a CTC head, thereby reducing the task complexity and achieving robust speaker invariance. This phoneme output then serves as the input to a fine-tuned Large Language Model (LLM), which reconstructs coherent words and sentences by leveraging broader linguistic context. Unlike existing methods that either predict words directly or rely on large-scale multimodal pre-training, our approach explicitly encodes intermediate linguistic structure while remaining highly data efficient. We demonstrate state-of-the-art performance on two challenging datasets, LRS2 and LRS3, where our method achieves significant reductions in Word Error Rate (WER) achieving a SOTA WER of 18.7 on LRS3 despite using 99.4$\%$ less labelled video data than the next best approach. Code is available here: \url{https://github.com/MarshallT-99/VALLR}
\end{abstract}    
\section{Introduction}
\begin{figure}[ht!]
\centering
\includegraphics[width=0.5\textwidth]{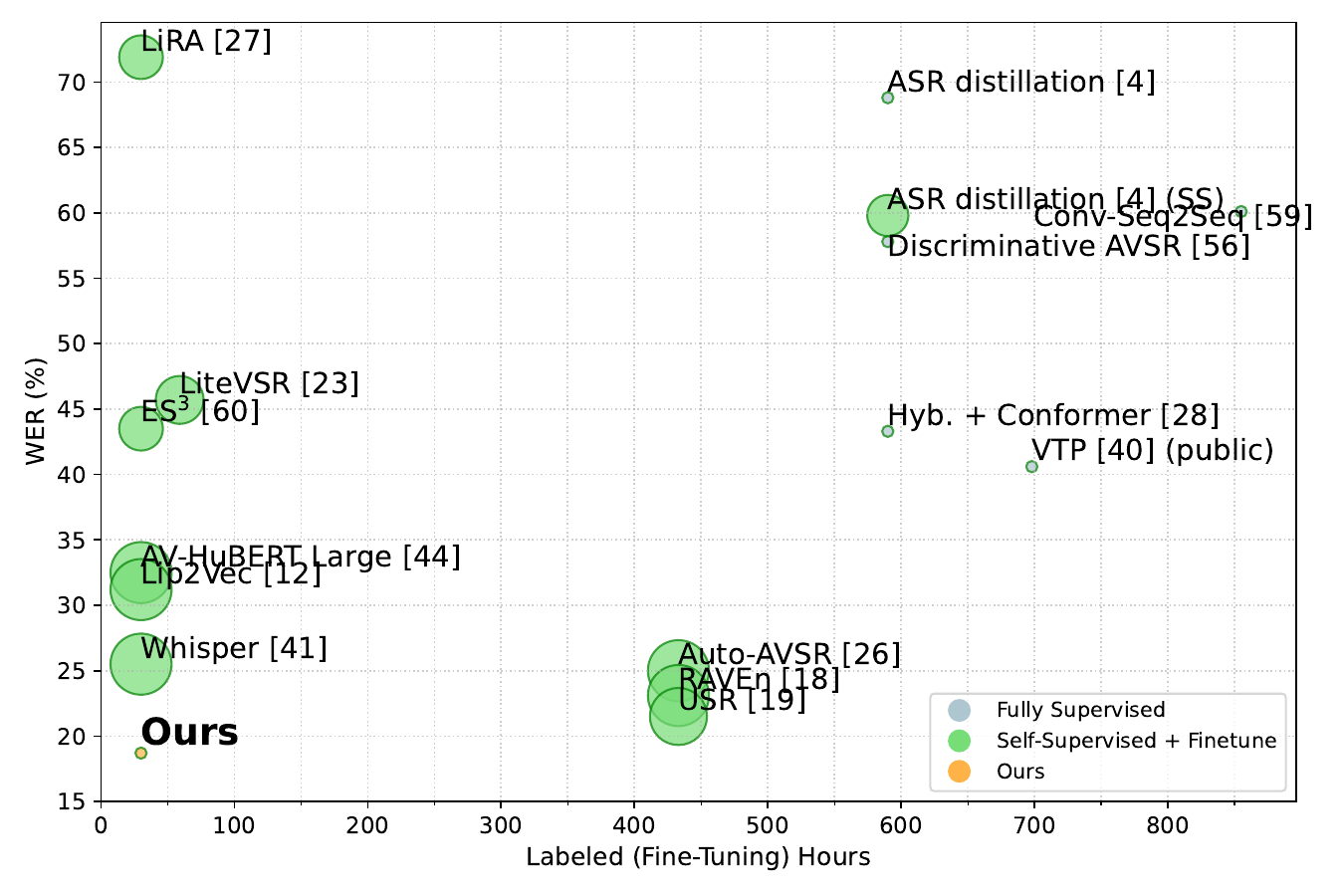}
\caption{Comparison between different models' performances in WER for Visual Automatic Speech Recognition on the LRS3 dataset~\cite{afouras2018lrs3} when compared with the amount of labelled training data. Circle size (green) denotes scale of pre-training data, while fine-tuned models (grey) are not pre-trained. Our model (orange) outperforms all existing approaches with just 30 hours of video training data, no self-supervised pre-training, and without the requirement for additional labeled visual data during fine-tuning.}
\label{fig:Comparison}  
\end{figure}

\begin{figure*}[ht!]
\centering
\includegraphics[width=\textwidth]{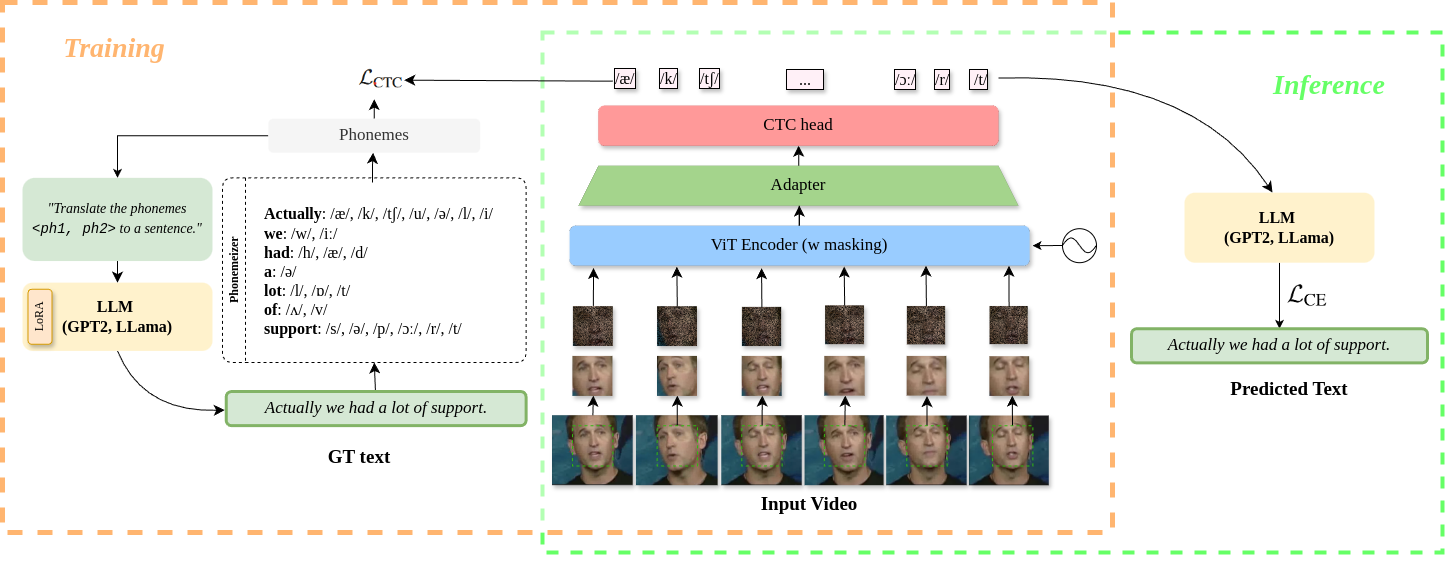}
\caption{An overview of our approach. First we extract facial regions from 16 frames of input video. We apply random pixel masking at 50\% probability, add positional embedding, and encode visual features via a vision transformer encoder (ViT-base \cite{1616}). We implement temporal downsampling via 1D convolution and then use a CTC linear head to predict sequences of phonemes. During training, we also fine-tune an LLM to reconstruct sentences from phonemes with a text-only dataset \cite{CMU}. During inference, the phonemes from the CTC head are processed via the LLM to reconstruct the predicted text. This can be performed end-to-end or in two stages, depending on available resources.}
\label{fig:Overview}
\end{figure*}

\noindent
Lip Reading, or Visual Automatic Speech Recognition (V-ASR), involves interpreting spoken language from visual cues such as lip movements and facial expressions. As a natural skill, humans use it to supplement auditory information, and as a technology, it has profound potential for enhancing accessibility, particularly for the Deaf and hard-of-hearing communities, as well as for applications in noisy or privacy-sensitive environments \cite{Lip-Reading-Survey}. Despite substantial advancements in ASR, the visual-only interpretation of speech remains an unresolved challenge due to inherent uncertainties in lip movements and their complex temporal dynamics. A central difficulty in V-ASR stems from the inherent ambiguity of visemes, the visual equivalents of phonemes, which often appear nearly identical for different sounds (e.g., ‘p’ and ‘b’)~\cite{bear2014phoneme, Lip-Reading-Survey}. Further complicating this task are coarticulation effects~\cite{viseme-problem}, where adjacent sounds blur one another’s articulation, making phoneme boundaries visually unclear. Although some recent methods have focussed on robust intermediary representations to deal with these cases, such as leveraging subword-level predictions \cite{subword}, or quantized latent representations \cite{Lip2Vec}, these approaches can still fail to capture broader contextual cues, limiting their ability to resolve visually similar phonemes that might otherwise be distinguishable through sentence-level semantics.

\par
\noindent One approach to this challenge is to leverage large language models (LLM's) and attempt to map raw lip movements directly to sentences, thus bypassing the issue of viseme ambiguity \cite{lipreadingllm, lipreadingllm2}. However, bridging the gap between high-dimensional, unstructured visual inputs and detailed textual representations is resource intensive and typically requires huge datasets, or specialized architectures. Moreover, these purely end-to-end pipelines often lack an interpretable intermediary step, making it unclear how the model is handling viseme ambiguity at a finer linguistic granularity. This can lead to hallucination effects, which become more critical in lip reading technologies designed for accessibility, or security applications. 

\par
\noindent Our two-stage, phoneme-centric method addresses both issues. In the first stage, we map short windows of video frames to a discrete, interpretable phoneme sequence. This intermediate representation is significantly easier to learn and more robust against speaker or environmental variations, as phonemes abstract away speaker-specific attributes. Subsequently, in the second stage, we fine-tune an LLM for the task of reconstructing sentences from phonemes. By segmenting the problem, we combine the interpretability and efficiency of phoneme-based prediction with the sophisticated contextual reasoning of modern LLMs. This \textit{phoneme$\rightarrow$sentence} pipeline also aligns with psycholinguistic models of speech perception, where phoneme-level processing naturally precedes lexical access \cite{cohort, word-recognition}.

\noindent Several advantages emerge from this design: 
\begin{itemize}[leftmargin=10pt]
    \item[(i)] \textbf{Compact Target Space:} Predicting only 38 phoneme classes avoids learning entire word-level vocabularies, making the model more data-efficient and simpler to train.
    \item[(ii)] \textbf{LLM-Enhanced Accuracy:} A large language model, pre-trained for \emph{phoneme}\(\to\)\emph{sentence} reconstruction, removes the requirement of direct word-level predictions from RGB data, reducing the need for complex multi-modal training.
    \item[(iii)] \textbf{Data Availability:} Exploiting easily available \emph{phoneme}\(\to\)\emph{sentence} text pairs eliminates reliance on extensive lip-reading video data for pre-training.
    \item[(iv)] \textbf{Interpretability \& Error Analysis:} The two-stage design yields an explicit phoneme-level output, constraining recognition errors to local phoneme segments, which the LLM subsequently repairs at the word level reducing errors at a sentence level.
\end{itemize}

\section{Related Work}

Here, we review key categories of approaches and highlight their advantages and limitations in the context of our proposed phoneme-centric, language-aware approach.

\par
\noindent \textbf{Lip Reading:} Early methods predominantly relied on statistical models such as Hidden Markov Models (HMMs) to aggregate temporal sequences of lip movements alongside hand-crafted feature extractors \cite{markov-viseme, markov-viseme-2, temporalseq, avasr, Lip-Reading-Survey, vsrreview}. With the advent of deep learning, coupled with the availability of large-scale datasets such as Grid \cite{The-Grid-Corpus}, LRW \cite{LRW}, LRS2 \cite{LRS2}, and LRS3 \cite{afouras2018lrs3}, significant progress has been made in recent years. Initially, Convolutional Neural Networks (CNNs) were applied to the task of word recognition \cite{LRW}, with additional temporal processing via RNN's \cite{RNN_T, lstmlip, LSTM_LIP}. With greater compute came the possibility to interpret full sentences with approaches applying Automatic Speech Recognition (ASR) methodologies to the Visual Automatic Speech Recognition (V-ASR) task. These included sequence-to-sequence models \cite{deepav}, CTC approaches \cite{lipnet,LargescaleAVSR}, and hybrid models \cite{hybrid, hybrid2}. These models improved recognition accuracy by learning hierarchical representations of visual features and modelling temporal dependencies. The application of transformer models \cite{ViT_3D, tong2022videomae, vaswani2017attention, 1616, Hubert} in V-ASR brought further performance gains in combination with extensive self-supervised pre-training. Methods such as AV-HuBERT \cite{AV_HuBERT} and LiteVSR \cite{LiteVSR} leveraged self-supervised learning to learn robust representations from large amounts of unlabelled video data.

\par
\noindent\textbf{AV-ASR vs V-ASR:} Lip reading approaches can be separated into Visual Automatic Speech Recognition (V-ASR) and Audio-Visual Automatic Speech Recognition (AV-ASR). In the AV-ASR task, labelled audio data is available at training time which can be used to guide the visual encoder in various ways. These can include generating synthetic additional pre-training data \cite{SynthVSR, Auto_AVSR}, or simply fusing audio-visual input features \cite{LargescaleAVSR,multi-modal, AV_HuBERT, ViT_3D}. Other methods have used knowledge distillation to inductively transfer knowledge from pre-trained ASR models to visual encoders \cite{ASR_distill, LiRA}. However, these methods depend on audio data or pre-trained ASR models, which may not be available or reliable in scenarios where lip reading is most needed, such as noisy environments. V-ASR, on the other hand, is a more challenging task that relies only on visual data in the training stage \cite{vsrreview, Lip-Reading-Survey, spatiotemporal, spatio-temporal2}. Recent methods have also shown the effectiveness of adapting visual inputs to pre-trained ASR models \cite{speechlipreading, Lip2Vec, LiteVSR} without the need for audio data during training. This method is particularly powerful since the ASR model already has the ability to reconstruct latent representations to words from the pre-training task, however it relies on large amounts of additional data to learn robust visual representations and map them to the audio latent space \cite{speechlipreading}. 

\noindent\textbf{Lip Reading with LLMs:} Recent approaches have explored visual attention mechanisms for sub-word units \cite{subword, PBLR, LVBVSRLR, stafylakis2018zero} to capture fine-grained linguistic features in lip reading. However, reconstructing these units into coherent sentences remains challenging for networks lacking robust contextual and semantic reasoning capabilities. Early phoneme-based methods \cite{el2023developing, thangthai2018comparing} similarly struggled with effective decoding to word-level representations and these methods were outpaced in performance by end-to-end approaches \cite{lipnet} which could leverage extensive data. As such they have not been fully explored in the context of modern architectures and techniques.

\par
\noindent For example, recent approaches which have integrated Large Language Models (LLMs) \cite{GPT, llama} into lip-reading pipelines \cite{lipreadingllm, lipreadingllm2} have started from mapping visual features directly to LLM text embeddings. However, by omitting explicit intermediate representations, these approaches inherit phonetic ambiguities that propagate through the network. This manifests as increased word error rates and reduced robustness, thus requiring substantially more training data to achieve generalization.

\par
\noindent Our work re-examines phonemes as an interpretable discrete representation bridging visual and linguistic domains. Rather than forcing LLMs to perform \textit{visual$\rightarrow$sentence} translation requiring expensive visual-text alignment – we constrain the LLM's role to \textit{phoneme$\rightarrow$sentence} reconstruction. This decomposition offers three key advantages: (1) Phoneme-to-text translation is a well-constrained task requiring only lightweight LLM fine-tuning using abundant textual corpora; (2) Speaker-independent phoneme representations eliminate the need for visual adaptation in the language model; (3) The modular architecture prevents error propagation between visual and linguistic domains while maintaining interpretability.

\section{Method}
\label{sec:method}

In this section, we present our two-stage \emph{phoneme-centric} approach to visual-only lip reading. Figure~\ref{fig:Overview} provides an overview of the framework. Our goal is to learn a function
\begin{equation}
\begin{aligned}
f:\quad \mathbf{X} &= \{x_1, x_2, \ldots, x_T\}\\
&\longmapsto \;\mathbf{Ph} = \{ph_1, ph_2, \ldots, ph_m\}\\
&\longmapsto \;\mathbf{S} = \{s_1, s_2, \ldots, s_M\}.
\end{aligned}
\end{equation}
where \(\mathbf{X} \in \mathbb{R}^{T \times H \times W \times 3}\) is a video sequence of \(T\) frames (with height \(H\) and width \(W\)), \(\mathbf{Ph}\) denotes the phoneme sequence, and \(\mathbf{S}\) is the reconstructed sentence. Specifically, we decompose the task into:
\begin{enumerate}[leftmargin=15pt]
    \item \textit{Video}\(\to\)\textit{Phoneme}: Map the sequence of video frames to phonemes using a Vision Transformer and CTC head.
    \item \textit{Phoneme}\(\to\)\textit{Sentence}: Convert phonemes to a coherent word sequence with a fine-tuned Large Language Model (LLM).
\end{enumerate}

\noindent The model comprises four primary components:

\begin{itemize}
    \item \textit{Visual Feature Extractor}: Captures relevant features from each frame via a Vision Transformer.
    \item \textit{Adapter Network with Temporal Downsampling}: Reduces sequence length to accommodate CTC alignment.
    \item \textit{CTC Head}: Predicts phoneme sequences without requiring explicit temporal labels.
    \item \textit{LLM for Phoneme-to-Sentence Reconstruction}: Translates predicted phonemes into final word sequences.
\end{itemize}

\subsection{Pre-Processing Pipeline}
\label{subsec:preprocess}
Each frame \(x_i\) is first passed through a face-detection model~\cite{MediaPipe} to identify and crop the speaker's face region, centered on the speaker's lips. The detected region is then cropped, resized to \(224 \times 224\), and normalized for batch processing.

\subsection{Video Transformer}
\label{subsec:video_transformer}
We employ a Vision Transformer (ViT)~\cite{ViT_3D} to encode the spatio-temporal information in the face region. Let \(\mathrm{ViT}(\cdot)\) denote this encoding function. The input \(\mathbf{X} \in \mathbb{R}^{T \times H \times W \times 3}\) is chunked into patches, embedded, and equipped with positional encodings before passing through multiple transformer blocks. We obtain:
\begin{equation}
\mathbf{Z} \;=\; \mathrm{ViT}\bigl(\mathbf{X}\bigr) \;\in\; \mathbb{R}^{T \times D},
\end{equation}
where \(D\) is the transformer output dimension per frame. During training, we randomly mask portions of the input patches to promote robust feature learning and improve generalization.

\subsection{Adapter Network with Temporal Downsampling}
\label{subsec:adapter}
The sequence length \(T\) of the ViT output can be prohibitively large for the subsequent CTC module. To reduce the temporal dimension, we apply a sequence of 1D convolutions and pooling layers, as shown in Table~\ref{tab:adapter}. Formally,
\begin{equation}
\mathbf{G}_{\text{down}} \;=\; \mathrm{Adapt}\bigl(\mathbf{Z}\bigr) \;\in\; \mathbb{R}^{T' \times C_{\text{adapter}}},
\end{equation}
where \(T' < T\) and \(C_{\text{adapter}}\) is an intermediate feature dimension aligned to the CTC head input.


\begin{table}[t]
\centering
\resizebox{0.5\textwidth}{!}{
\begin{tabular}{c l l c c c c l}
\toprule
Stage & Module & In \(\to\) Out Channels & Kernel & Stride & Padding & Downsample Factor & Output Length \\
\midrule
1 & Conv1d & feature\_size \(\to\) adapter\_dim & 5 & 2 & 2 & \(\times2\) & \(L/2\) \\
      & BatchNorm1d & adapter\_dim \(\to\) adapter\_dim & – & – & – & & \\
      & ReLU & — & – & – & – & & \\
\midrule
2 & Conv1d & adapter\_dim \(\to\) adapter\_dim & 3 & 2 & 1 & \(\times2\) & \(L/4\) \\
& BatchNorm1d & adapter\_dim \(\to\) adapter\_dim & – & – & – & & \\
& ReLU & — & – & – & – & & \\
\midrule
3 & Conv1d & adapter\_dim \(\to\) adapter\_dim & 3 & 2 & 1 & \(\times2\) & \(L/8\) \\
& BatchNorm1d & adapter\_dim \(\to\) adapter\_dim & – & – & – & & \\
& ReLU & — & – & – & – & & \\
\midrule
4 & Conv1d & adapter\_dim \(\to\) adapter\_dim & 3 & 3 & 1 & \(\times3\) & \(L/24\) \\
& BatchNorm1d & adapter\_dim \(\to\) adapter\_dim & – & – & – & & \\
& ReLU & — & – & – & – & & \\
\midrule
5 & AvgPool1d & adapter\_dim \(\to\) adapter\_dim & 5 & 8 & 0 & \(\times8\) & \(L/192\) \\
\midrule
6 & Linear & adapter\_dim \(\to\) CTC\_dim & - & - & - & \(\times1\) & \(L/192\) \\
& ReLU & — & – & – & – & & \\
\bottomrule
\end{tabular}
}
\caption{Architecture of the adapter for temporal downsampling, where the input length is \(T=1568\) and final output length is \(T=8\).}
\end{table}

\subsection{CTC Head}
\label{subsec:ctc}

An MLP with one hidden layer transforms 
\(\mathbf{G}_{\text{down}} \in \mathbb{R}^{T' \times C_{\text{adapter}}}\) 
into phoneme logits:
\begin{equation}
\label{eq:ctc_logits}
\mathbf{H}_{\text{ctc}} \;=\; \mathrm{MLP}\bigl(\mathbf{G}_{\text{down}}\bigr) 
\;\in\; \mathbb{R}^{T' \times N_{\mathrm{phonemes}}},
\end{equation}
where \(N_{\mathrm{phonemes}}\) is the size of our phoneme set (39 English phonemes plus a blank symbol). We then apply a logarithmic softmax along the phoneme dimension to obtain log probabilities 
\(\log p_{\mathrm{ctc}}(ph_t \mid \mathbf{X})\) at each time step. 

Since lip movements do not align cleanly with discrete phoneme boundaries, we adopt the Connectionist Temporal Classification (CTC) loss~\cite{CTC}:
\begin{equation}
\label{eq:ctc_loss}
\mathcal{L}_{\text{CTC}} \;=\; 
-\ln \Bigl(\sum_{\alpha \in \mathcal{A}(\mathbf{Ph})} 
P_{\mathrm{ctc}}\bigl(\alpha \mid \mathbf{X}\bigr)\Bigr),
\end{equation}
where \(\mathbf{Ph} = (ph_1,\ldots,ph_m)\) is the ground-truth phoneme sequence, 
\(\mathcal{A}(\mathbf{Ph})\) is the set of all valid alignments, and 
\(P_{\mathrm{ctc}}(\alpha \mid \mathbf{X})\) is the probability of alignment \(\alpha\). 
Minimizing \(\mathcal{L}_{\text{CTC}}\) enables the model to learn frame-to-phoneme mappings without explicit per-frame labels. At inference, we perform beam search over the CTC log probabilities to decode the final phoneme sequence.

\subsection{Phoneme-to-Sentence Reconstruction Using LLM}
\label{subsec:phoneme2sentence}
The predicted phoneme sequence \(\widehat{\mathbf{Ph}}\) is derived by decoding the CTC logits from Eq.~\eqref{eq:ctc_logits} via beam search. We then feed \(\widehat{\mathbf{Ph}}\) into a Large Language Model (\(\mathrm{LLM}\)) that is fine-tuned to map phonemes to sentences:
\begin{equation}
\label{eq:llm_mapping}
\mathbf{S} \;=\; \mathrm{LLM}\bigl(\widehat{\mathbf{Ph}}\bigr) \;=\; (s_1, s_2, \ldots, s_M).
\end{equation}

\paragraph{LoRA Fine-Tuning.}
To efficiently adapt the LLM, we employ Low-Rank Adaptation (LoRA) \cite{hu2022lora} , injecting two low-rank matrices \(\mathbf{A}\) and \(\mathbf{B}\) into each linear layer. This greatly reduces the number of trainable parameters compared to full-model fine-tuning. Specifically, for a linear layer \(\mathbf{W}_{\text{orig}} \in \mathbb{R}^{d \times d}\):
\begin{equation}
\label{eq:lora}
\mathbf{W}' \;=\; \mathbf{W}_{\text{orig}} + \mathbf{A}\,\mathbf{B}, \quad 
\mathbf{A} \in \mathbb{R}^{d \times r}, \;
\mathbf{B} \in \mathbb{R}^{r \times d}, 
\end{equation}
where \(r \ll d\). Only \(\mathbf{A}\) and \(\mathbf{B}\) are updated during training.

\paragraph{Pretraining LLM Objective.}
We fine-tune the LLM on a large \textit{phoneme$\to$sentence} corpus generated from WikiText~\cite{WikiText}. Let \(\mathbf{S} = (s_1, \ldots, s_M)\) be the ground-truth sentence for a phoneme sequence \(\mathbf{Ph}\). We employ the cross-entropy loss:
\begin{equation}
\label{eq:llm_ce_loss}
\mathcal{L}_{\text{CE}} \;=\; -\sum_{t=1}^{M} \ln\, p\bigl(s_t \,\big\vert\, \mathbf{Ph},\, s_{1:t-1}\bigr),
\end{equation}
where \(p(s_t \mid \mathbf{Ph}, s_{1:t-1})\) is the likelihood of predicting the correct word \(s_t\), given the phoneme sequence and previously generated words. During inference, we prompt the LLM with \(\widehat{\mathbf{Ph}}\) to generate \(\widehat{\mathbf{S}}\).

\subsection{Training Procedure}
\label{subsec:training}
\noindent \textbf{Stage 1: Video\(\to\)Phoneme.} We train the visual extractor, adapter, and CTC head to minimize \(\mathcal{L}_{\text{CTC}}\) (Eq.~\ref{eq:ctc_loss}) on video-phoneme pairs. By learning without strict alignment constraints, the model remains flexible to a wide range of speaking speeds and styles.

\vspace{4pt}
\noindent \textbf{Stage 2: Phoneme\(\to\)Sentence.} We independently fine-tune the LLM via cross-entropy loss (Eq.~\ref{eq:llm_ce_loss}) on text-based phoneme$\to$sentence datasets. This leverages easily obtained text data, allowing the LLM to learn phonetic-linguistic mappings separately from visual modeling. 

\vspace{4pt}
\noindent \textbf{Inference.} We compose the trained modules to form:
\[
\mathbf{X} \;\xrightarrow{f_{\mathrm{vis}}}\; \widehat{\mathbf{Ph}} 
\;\xrightarrow{f_{\mathrm{LLM}}}\; \widehat{\mathbf{S}},
\]
where \(\widehat{\mathbf{Ph}}\) is decoded from CTC logits, and \(\widehat{\mathbf{S}}\) is the final sentence. This approach is highly modular and data-efficient, as each stage (visual and linguistic) is learned with its own objective and dataset, yet easily integrated at inference time.

\section{Experiments}
This section outlines the datasets and pre-processing methods used for training the model architecture.
\subsection{Data}
The visual feature extractor, the adapter with downsampling, and CTC head are trained on the LRS2 \cite{LRS2} and LRS3 \cite{afouras2018lrs3} datasets.
The LLM is fine-tuned on the WikiText \cite{WikiText} dataset. 

\noindent \textbf{LRS2 \cite{LRS2}:} The LRS2 dataset consists of 144,482 video clips of spoken sentences from BBC television, consisting of approximately 224.5 hours of footage and each sentences is up to 100 characters in length. The videos are divided into a pre-training set with 96,318 utterances (195 hours), a training set with 45,839 utterances (28 hours), a validation set with 1,082 utterances (0.6 hours) and a test set with 1,243 utterances (0.5 hours). 

\noindent \textbf{LRS3 \cite{afouras2018lrs3}:} The LRS3 dataset describes the largest public audio-visual English dataset collected consists of clips from over 5,000 TED and TEDx talks totaling 438.9 hours. It contains 438.9 hours with 151,819 utterances. Specifically, there are 118,516 utterances in the ‘pre-train’ set (408 hours), 31,982 utterances in the ‘train-val’ set (30 hours) and 1,321 utterances in the ‘test’ set (0.9 hours).

\noindent
In our setting, unlike other approaches, we do not utilise the pre-training set and train our model on only the 28-hour and 30-hour partitions.

\noindent \textbf{Phoneme Sentence Pairs.} Finally, the dataset used for training the LLM was the WikiText \cite{WikiText} dataset with sentences converted into lists of phonemes using the CMU dictionary \cite{CMU}. These phonemes were then masked and paired with the original sentences. 

\subsection{Evaluation Metrics}
Following previous works \cite{LRW, LRS2} we adopt Word Error Rate \cite{WER} (WER) as our evaluation metric for Lip-Reading. WER calculates the percentage of errors in the predicted text compared to the ground truth, accounting for substitutions, insertions, and deletions. Similar to previous studies \cite{LRW, LRS2,afouras2018lrs3, subword, speechlipreading}, we report results on all recent AV-ASR and V-ASR approaches.

\definecolor{lightgray}{gray}{0.9} 

\section{Results}
In this section, we present the results of our method and compare them against those of existing SOTA approaches. 

\noindent \textbf{Public vs Private Data on LRS3 \cite{afouras2018lrs3}:} In Table 1, we compare different V-ASR and AV-ASR models based on their WER on the LRS3 dataset, we split the models into two categories; Fully Supervised models with publicly available data and models trained on large-scale non-publicly available datasets.

\noindent The Fully Supervised Models rely solely on publicly available labelled datasets with the best WER being 40.6\%, which was achieved by VTP \cite{subword}. The non-public dataset models leverage extensive, proprietary datasets, achieving much lower WERs with the best being 12.5\% by LP \cite{LP}.
\begin{table}[h!]
\centering
\resizebox{0.5\textwidth}{!}{ 
\begin{tabular}{lccc}
\toprule
\textbf{Method} & \textbf{Unlabeled (hrs)} & \textbf{Labeled (hrs)} & \textbf{WER (\%)} \\
\midrule
\rowcolor{lightgray} \multicolumn{4}{l}{\textit{Fully supervised models with publicly available data}} \\
ASR distillation \cite{ASR_distill} & - & 590 & 68.8 \\
Conv-Seq2Seq \cite{spatiotemporal} & - & 855 & 60.1 \\
Discriminative AVSR \cite{multi-modal} & - & 590 & 57.8 \\
Hyb. + Conformer \cite{conformer} & - & 590 & 43.3 \\
VTP \cite{subword} & - & 698 & 40.6 \\
\midrule
\rowcolor{lightgray} \multicolumn{4}{l}{\textit{Trained on large-scale non-publicly available datasets}} \\
Deep-AV-SR \cite{deepav} & - & 1,519 & 58.9 \\
Large-scale AV-SR \cite{LargescaleAVSR} & - & 3,886 & 55.1 \\
RNN-T \cite{RNN_T} & - & 31,000 & 33.6 \\
VTP \cite{subword} & - & 2,676 & 30.7 \\
ViT-3D \cite{ViT_3D} & - & 90,000 & 17.0 \\
LP \cite{LP} & - & 1,000,000 & 12.5 \\
\midrule
\rowcolor{lightgray} \multicolumn{4}{l}{\textit{Self-supervised pre-training + Supervised fine-tuning on LRS3}} \\
LiRA \cite{LiRA} & 433 & 30 & 71.9 \\
ASR distillation \cite{ASR_distill} & 334 & 590 & 59.8 \\
LiteVSR \cite{LiteVSR} & 639 & 59 & 45.7 \\
ES$^3$ \cite{zhang2024es3} & 433 & 30 & 43.5 \\
AV-HuBERT Large \cite{AV_HuBERT} & 1,759 & 30 & 32.5 \\
Lip2Vec \cite{Lip2Vec} & 1,759 & 30 & 31.2 \\
Sub-Word\cite{subword} & 2,676 & 2,676 & 30.7 \\
SynthVSR \cite{SynthVSR} & 3,652 & 438 & 27.9 \\
Whisper \cite{speechlipreading} & 1,759 & 30 & 25.5 \\
Auto-AVSR \cite{Auto_AVSR} & 1,759 & 433 & 25.0 \\
LLaMA-AVSR \cite{LLaMA-AVSR} & - & 1756 & 24.0 \\
RAVEn \cite{raven} & 1759 & 433 & 23.1 \\
USR \cite{unispeech} & 1,326 & 433 & 21.5 \\
\midrule
\rowcolor{lightgray} \multicolumn{4}{l}{\textit{Supervised finetuning only on LRS3}} \\
\textbf{Ours} & \textbf{-} & \textbf{30} & \textbf{18.7} \\
\bottomrule
\end{tabular}
}
\caption{Comparison of Word Error Rate (WER) across different models for visual-only speech recognition on the LRS3 dataset. The table is divided into three sections: fully supervised models trained on publicly available data, models trained on large-scale non-public datasets, and models that leverage self-supervised pre-training with supervised fine-tuning on LRS3 \cite{afouras2018lrs3}. Each entry details the amount of labelled and unlabelled data used and the resulting WER. Our approach, tested with various language models, achieves the lowest WER of 22.1\% with the Llama 3.2-3B model, demonstrating the effectiveness of our phoneme-based, LLM-assisted approach.}
\end{table}

\begin{table}[h!]
\centering
\resizebox{0.5\textwidth}{!}{
\begin{tabular}{lccc}
\toprule
\textbf{Method} & \textbf{Unlabeled (hrs)} & \textbf{Labeled (hrs)} & \textbf{WER (\%)} \\
\toprule
\rowcolor{lightgray} \multicolumn{4}{l}{\textit{Fully supervised models with publicly available data}} \\
Auto-AV-SR \cite{Auto_AVSR} & - & 818 & 27.9 \\
Auto-AV-SR \cite{Auto_AVSR} & - & 3448 & 14.6 \\
\midrule
\rowcolor{lightgray} \multicolumn{4}{l}{\textit{Self-supervised pre-training + Supervised finetuning on LRS2 for AVSR}} \\
LiRA \cite{LiRA} & 1,759 & 433 & 38.8 \\
ES$^3$ \cite{zhang2024es3} & 1,759 & 223 & 26.7 \\
Sub-Word \cite{subword} & 2,676 & 2,676 & 22.6 \\
RAVEn \cite{raven} & 1,759 & 223 & 17.9 \\
USR \cite{unispeech} & 1,759 & 223 & 15.4 \\
\midrule
\rowcolor{lightgray} \multicolumn{4}{l}{\textit{Supervised finetuning only on LRS2}} \\
\textbf{Ours} & \textbf{-} & \textbf{28} & \textbf{20.8} \\
\bottomrule
\end{tabular}
}
\caption{Comparison of Word Error Rate (WER) for various self-supervised pre-training and supervised fine-tuning methods evaluated on the LRS2 \cite{LRS2} dataset. Our method achieves competitive performance with other methods, without the requirement of extensive pre-training, additional labelled data, or additional audio modality as in the best performing approaches. We are the only method to train on only the LRS2 train dataset.}
\end{table}

\noindent \textbf{LRS3 \cite{afouras2018lrs3}:} In Table 1 we compare our results against other methods based on results from the LRS3 dataset. The results in Table 1 demonstrate a consistent trend: models incorporating extensive self-supervised pre-training followed by fine-tuning achieve lower WERs than those using only supervised approaches. However, our model achieves the lowest WER on the LRS3 dataset at 18.7\% without self-supervised pre-training or additional fine-tuning data.

\noindent \textbf{LRS2 \cite{LRS2}:} In Table 2, we compare our method against several other state-of-the-art approaches for visual-only lip reading, focusing on Word Error Rate (WER) on the LRS2 dataset. From Table 2 we can see that increasing the amount of data available helps improve a model's WER. However, we can also see that even without increasing the size of the dataset, our model still outperforms other models that only use publicly available data, achieving the lowest WER at 20.8\%.

\begin{figure}[h!]
\centering
\includegraphics[width=0.5\textwidth]{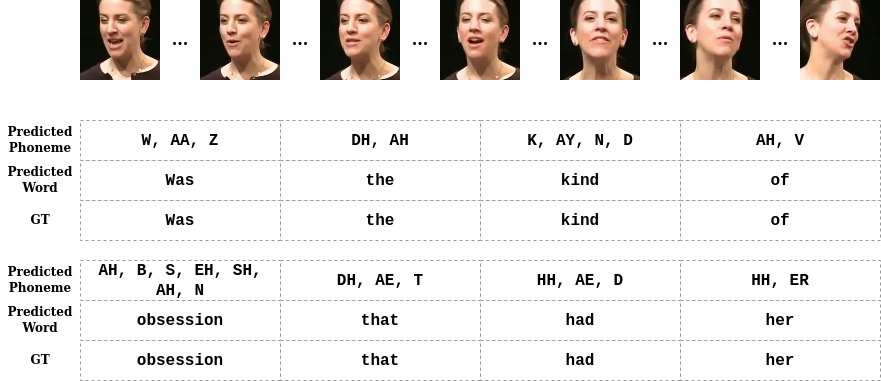}
\caption{Example of the model's phonetic and sentence outputs from a sample in the LRS3 dataset \cite{afouras2018lrs3}. The table illustrates the model’s ability to predict a sequence of phonemes from visual input, which are then reconstructed into a coherent sentence by the LLM. In this example all of the phonemes are predicted correctly and the words are recreated correctly.}
\label{fig:correct_sentences}  
\end{figure}

\begin{figure}[h!]
\centering
\includegraphics[width=0.5\textwidth]{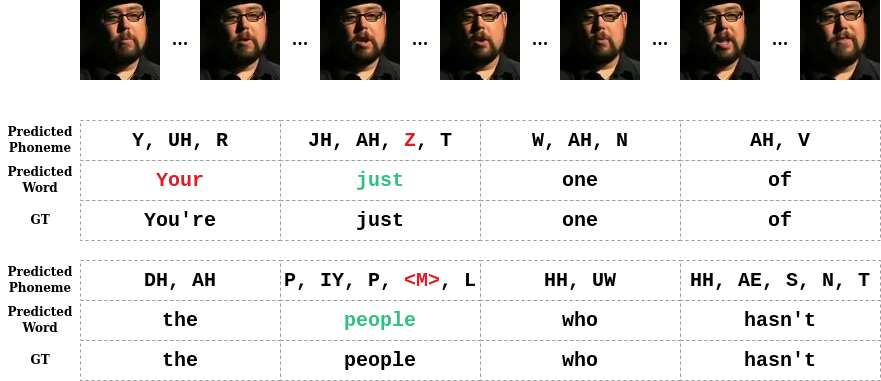}
\caption{Comparison of the model's phonetic and sentence outputs with ground truth from a sample in the LRS3 \cite{afouras2018lrs3} dataset. Red shows an incorrect prediction, \textcolor{red}{\textbf{M}} shows a missing prediction and green shows a correction from phonemes to words. In this example, even though the model incorrectly predicts certain phonemes, the LLM can correctly recreate the word but struggles to recreate homophones. }
\label{fig:corrected_sentences}  
\end{figure}

\noindent Therefore as shown in Tables 2 and 3, we can see that our method provides a lower WER with the same amount, or less data, without the requirement of self-supervised pre-training as current approaches \cite{ASR_distill, LiRA, LiteVSR, AV_HuBERT, SynthVSR, Auto_AVSR, Lip2Vec, speechlipreading}. This large increase in performance and generalisation can be explained by the powerful combination of the multi-task objectives and the ability of the LLM to reconstruct sentences accuratley even when phonemes are predicted incorrectly. In Fig.~\ref{fig:corrected_sentences}, we find that the phonetic output of the visual encoder is sometimes incorrect or misses phonemes, but the LLM is able to still reconstruct the correct word (as in the example z$\rightarrow$s). We also show how the model still struggles with tricky homophones such as \textbf{\textit{your}} and \textbf{\textit{you're}}. In Fig.~\ref{fig:correct_sentences}, we show an additional example of the model's outputs at both phonetic and word levels, demonstrating how phonemes are correctly predicted and combined by the LLM for sentence reconstruction. For further examples please see the appendix.

\subsection{Ablations} In this section, we aim to understand the performance contribution of each component in the network and their performance on \textit{visual $\rightarrow$ phoneme} prediction and \textit{phoneme $\rightarrow$ word} reconstruction, alongside ablations related to architectural decisions, and the limitations of the approach.

\begin{figure}[h!]
\centering
\includegraphics[width=0.5\textwidth]{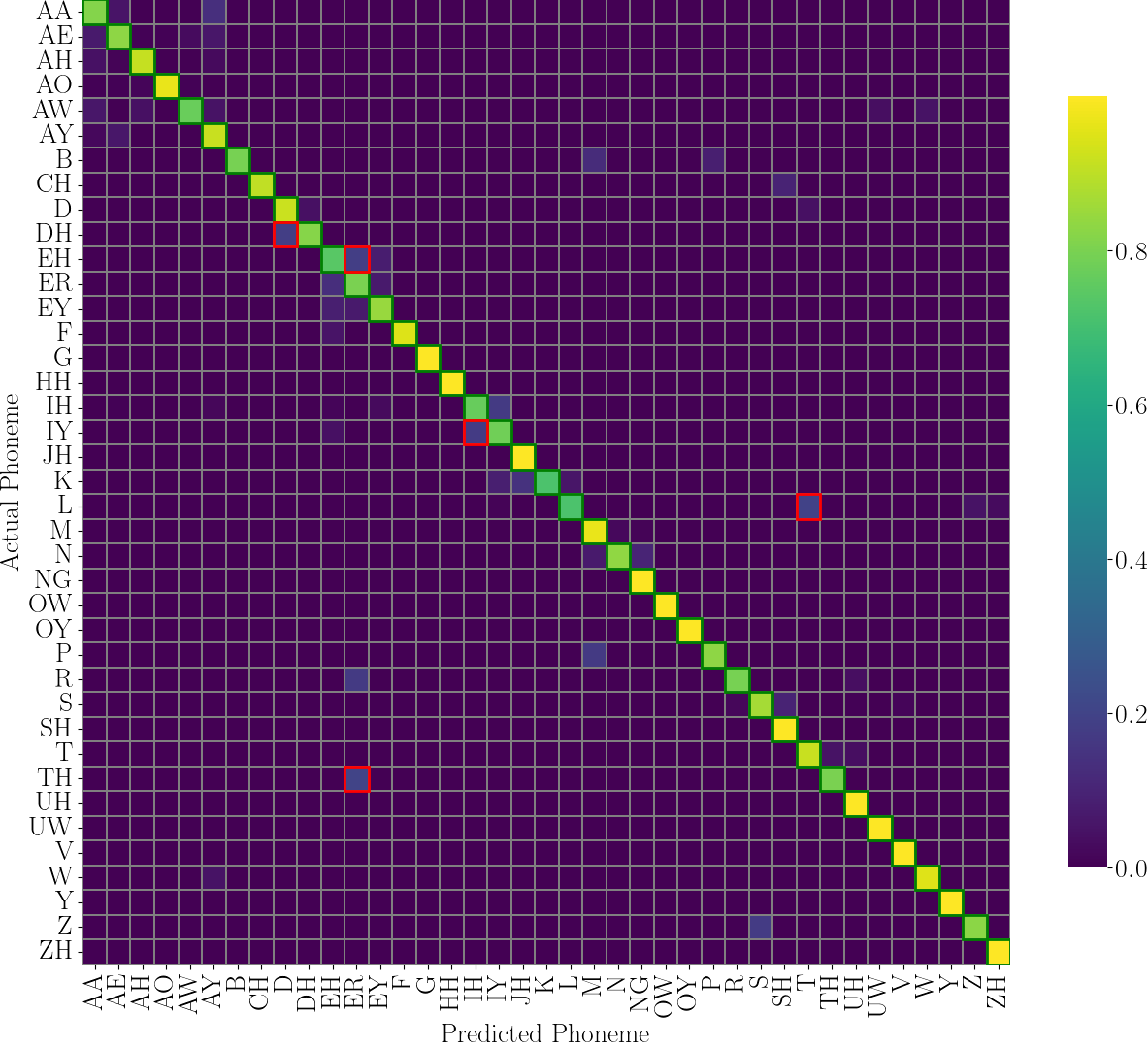}
\caption{Confusion matrix showing the performance on isolated phonemes of the LRS3 \cite{afouras2018lrs3} dataset. We observe a very high match rate between the predicted phonemes and the ground truth. In red, we show the most difficult phonemes for our model to identify. }
\label{fig:confusionmatrix}  
\end{figure}

\noindent \textbf{\textit{Visual $\rightarrow$ Phoneme}}: The confusion matrix in Fig.~\ref{fig:confusionmatrix} shows phoneme prediction performance of the model on the LRS3 \cite{afouras2018lrs3} dataset. The matrix indicates that some phonemes are misclassified more frequently than others, as seen by the off-diagonal elements, with the worst missed classifications being highlighted with a red bounding box. This bias could be due to the similarity in the visual representation of these sounds, making it challenging for the model to distinguish between them. For instance, /S/ and /Z/ may often be confused due to similar lip movements, leading to miss classifications between these phonemes.

\noindent \textbf{\textit{Phoneme $\rightarrow$ Word}}: In Table 4 we take the top 5 words in alphabetical order that are misclassified when predicting words directly from phonemes after the LLM pretraining stage. We can observe that these are words that may occur infrequently in the training set and are missclassified due to some phonetic similarities between the words.  

\begin{table}[h!]
\centering
\resizebox{0.5\textwidth}{!}{
\begin{tabular}{ccc}
\hline
\textbf{Phonemes} & \textbf{True Word} & \textbf{Predicted Word} \\
\hline
/EH/ /R/ /AH/ /N/ /S/ /AH/ /N/ & Aaronson & Erasmus \\
/AA/ /B/ /ER/ /G/ & Aaberg & Braggartly \\
/AH/ /S/ /EH/ /R /AH/ & Acerra & Esraa \\
/AE/ /S/ /AH/ /T/ /EY /T/ & Acetate & Asquette \\
/AH/ /K/ /AH/ /S/ /T/ /AH/ /M/ /D/ & Accustomed & Accosted \\
\hline
\end{tabular}
}
\caption{Examples of misclassification's between predicted words and ground truth based on phoneme sequences. Each row presents a sequence of phonemes, the intended (true) word, and the model's predicted word. Differences in predictions illustrate common misclassifications, often due to phonetic similarities between visually indistinct sounds.}
\label{tab:phoneme_comparison}
\end{table}

\noindent \textbf{Different LLMs:} In this ablation study, we evaluate three different Language models on our model architecture to understand the impact on word error rate (WER) performance. The first model, GPT 2 \cite{GPT}, serves as our baseline, consisting of the default GPT 2 \cite{GPT} without fine-tuning on lip-reading datasets. This model achieves a WER of 23.9\%, indicating the core components' foundational capabilities. The second model, Llama 3.2-1B \cite{llama} introduces fine-tuning on phoneme word pairs, enhancing the model’s contextual understanding of phoneme sequences. With this addition, Llama 3.2-1B \cite{llama} has a substantial improvement, reducing the WER to 22.8\%. Finally, Llama 3.2-3B \cite{llama} also incorporates fine-tuning for phoneme-to-word reconstruction. However, Llama 3.2-3B \cite{llama} is a larger model with more parameters than the Llama 3.2-1B \cite{llama} model, further lowering the WER to 18.7\%. This progression highlights the significant contributions of fine-tuning for purpose-specific LLMs. 

\begin{table}[h!]
\centering
\begin{tabular}{lcc}
\toprule
\textbf{Model} & \textbf{Parameters (B)} & \textbf{WER (\%)} \\
\midrule
GPT-2 Small \cite{GPT} & 0.12 & 33.9 \\
Llama 3.2-1B \cite{llama} & 1.0 & 22.8 \\
\textbf{Llama 3.2-3B \cite{llama}} & \textbf{3.0} & \textbf{18.7} \\
\bottomrule
\end{tabular}
\caption{Comparison of Word Error Rate (WER) for different large language models (LLMs) tested with our model. The table includes each model's parameter count (in billions) and resulting WER on the LRS3 \cite{afouras2018lrs3} dataset. GPT-2 \cite{GPT} Small with 0.12B parameters has the highest WER at 23.9\%. In contrast, Llama 3.2-3B \cite{llama} achieves the best WER of 18.7\%, highlighting that increased model capacity improves phoneme-to-word reconstruction accuracy.}
\end{table}

\noindent \textbf{Freezing the CTC Head:} Inspired by works that directly map visual features to pretrained ASR networks \cite{speechlipreading} we investigate the effect of initialising the CTC head in our model with weights fro  m a pre-trained Wav2Vec2 ASR model \cite{wav2vec2}. The CTC head plays a crucial role in mapping visual features to the phoneme sequences, and we hypothesized that initialising and freezing its parameters could reduce computational overhead while preserving feature alignment from audio pre-training. To test this hypothesis a baseline model with a non-frozen CTC head and a comparison model with a frozen CTC head were trained and a comparison of the results was made as shown in Table 6. Freezing the CTC head led to a slight increase in WER of 0.4\%, indicating a decrease in performance. This result suggests that while the frozen CTC head still provided a foundation for phoneme alignment, we did not require extensive audio pre-training to obtain superior performance.

\begin{figure}[h!]
\centering
\includegraphics[width=0.5\textwidth]{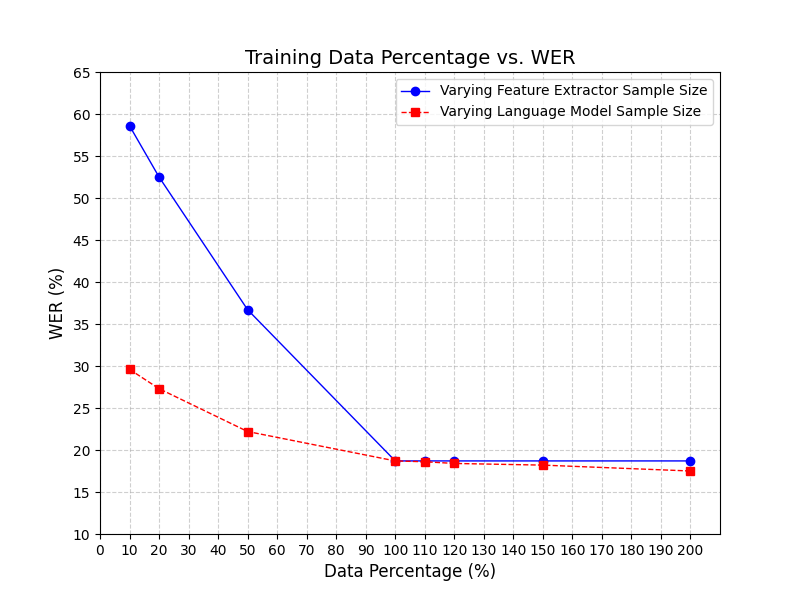}
\caption{Graph showing the effect of varying the sample size of the training datasets for both the Video $\rightarrow$ Phoneme network (in blue) and for the Phoneme $\rightarrow$ Sentence LLM (in red). The graph shows that decreasing the amount of data negatively impacts the model for both networks, increasing the amount of training data for the Video $\rightarrow$ Phoneme network has negligible effect and increasing the amount of training data for the Phoneme $\rightarrow$ Sentence LLM has a noticeable positive effect.} 
\label{fig:samlesize}  
\end{figure}

\noindent \textbf{Varying Sample Sizes:} To understand the impact of additional data on performance, we vary the quantity of data used to train both the \textit{Video$\rightarrow$Phoneme} network and the \textit{Phoneme$\rightarrow$Sentence} LLM and investigate the effect this has on the WER when evaluated on the LRS3 \cite{afouras2018lrs3} dataset. By varying the percentages of data used in training, we can see the effect on the model in Figure \ref{fig:samlesize}. Decreasing the amount of data used to train both models has the expected effect on the WER by increasing it. However, increasing the amount of training data for the Feature Extractor has a negligible effect after 30 hours of data. However, by increasing the amount of training data for the LLM we can also increase the effectiveness of the model and decrease the total WER down further to 17.5\%. The additional data is randomly sampled from both AvSpeech \cite{AVSepech} for the visual encoder, and a BookCorpus \cite{zhu2015aligning} replica to match the quantity of data in WikiText.

\noindent \textbf{End-To-End:} For our final study, we investigate the effect of removing the \textit{video $\rightarrow$ phoneme} stage and train the model to directly predict sentences from the video features, reproducing the method in \cite{subword} but with the same ViT encoder \cite{ViT_3D}, LLama LLM \cite{llama}, and limited data set. As shown in Table 6 we achieve a WER of 25.6\% on LRS3, demonstrating that the more parameterised transformer encoder improves performance over the CNN encoder used in \cite{subword}. However, our two stage approach still performs significantly better.

\begin{table}[h!]
\centering
\begin{tabular}{lcc}
\toprule
\textbf{Model} & \textbf{WER (\%)} \\
\midrule
Ours (Two Stage) & 18.7 \\
ViT + LLM (End to End) & 25.6 \\
Sub-Word (End to End) \cite{subword} & 30.7 \\

\bottomrule
\end{tabular}
\caption{Comparison of Word Error Rate (WER) between using the intermediary Phonemes and skipping the CTC head. Removing the intermediary representation increases the WER from 18.7 to 25.6, showing the importance of the intermediary step and the effectiveness of the phoneme centric fine-tuning.}
\end{table}

\section{Conclusion and Limitations}
\label{sec:conclusion}

We have introduced a two-stage, phoneme-centric framework for visual-only lip reading that first predicts phonemes from lip movements and then reconstructs coherent sentences via a fine-tuned LLM. This design sharply reduces word error rates on benchmark datasets (LRS2~\cite{LRS2} and LRS3~\cite{afouras2018lrs3}), providing improvements of 3.5\% and 6.8\% WER, respectively, over existing state-of-the-art approaches that focus on end-to-end connectionist approaches.

\noindent Despite these gains, the approach remains limited by the visual ambiguity of certain phonemes. Lip movements alone cannot always differentiate minimal pairs such as /s/ vs. /z/, and words that share near-identical phoneme sequences (e.g., ``Hello'' vs.\ ``Hallow'') can still result in misclassification. We plan to address these issues by refining the phoneme-to-word generation process, potentially through deeper linguistic context modeling and more specialized phoneme embeddings. Future work may also explore speaker-adaptive fine-tuning or multi-frame alignment strategies to better handle subtle visual distinctions. 

\section*{Acknowledgements}
This work was supported by the SNSF project ‘SMILE II’ (CRSII5 193686), the Innosuisse IICT Flagship (PFFS-21-47), EPSRC grant APP24554 (SignGPT-EP/Z535370/1) and through funding from Google.org via the AI for Global Goals scheme. This work reflects only the author’s views and the funders are not responsible for any use that may be made of the information it contains.
{
    \small
    \bibliographystyle{ieeenat_fullname}
    \bibliography{main}
}

\setcounter{page}{1}
\maketitlesupplementary
\section{Introduction}
\label{sec:introduction}
This supplementary material provides additional qualitative results from our proposed network architecture, evaluated on the LRS3 dataset \cite{afouras2018lrs3}. We include video examples referenced as {\ttfamily exampletranslation1.mp4} and {\ttfamily exampletranslation2.mp4} to demonstrate the effectiveness of the method in performing silent video captioning. These examples overlay the predicted captions on the original videos to offer an intuitive understanding of the predictions.
\noindent
Furthermore, we provide detailed analysis of common errors in phoneme-to-word reconstruction to identify limitations and strengths of the model. The additional results in this document are organized in three parts:

\noindent
\textbf{Tab.~\ref{supptab1}:} Common errors in phoneme $\rightarrow$ word predictions, such as homophones ({\ttfamily too} vs.\ {\ttfamily to}) or under-represented words ({\ttfamily sunflower} misinterpreted as {\ttfamily son} and {\ttfamily flower})

\noindent
\textbf{Tab.~\ref{suptab2}:} Challenging cases, including unseen names (e.g., {\ttfamily Kofi Annan}), demonstrating areas for future improvement with larger-scale pretraining.

\noindent
\textbf{Tab.~\ref{supptab3}:} Correct phoneme and word predicitions.

\noindent
These results complement the main paper and showcase the robustness of our approach while highlighting areas for refinement.

\section{Qualitative Results}
\label{sec:results}

In this section, we provide additional qualitative results in two parts. Errors in the phoneme-to-word reconstruction process are highlighted in \textcolor{red}{red} to draw attention to specific areas where the network requires improvement.

\subsection{Part 1: Common Errors}
In Tab.~\ref{supptab1}, we observe that common errors include substitutions of homophones like {\ttfamily too} and {\ttfamily to}, or {\ttfamily there} and {\ttfamily their}. Similarly, in examples involving the word {\ttfamily sunflower}, the model predicts {\ttfamily son} and {\ttfamily flower} as separate words. These errors are likely due to the absence of the compound word {\ttfamily sunflower} in the fine-tuning dataset, while the individual words {\ttfamily son} and {\ttfamily flower} are well-represented. Despite these mistakes, the resulting text remains logical and understandable.

\begin{table*}[t!]
\centering
\begin{tabular}{|c|p{5cm}|p{5cm}|p{5cm}|}
\hline
\textbf{File} & \textbf{Video $\rightarrow$ Phoneme (Errors in \textcolor{red}{red})} & \textbf{Phoneme $\rightarrow$ Word (Errors in \textcolor{red}{red})} & \textbf{Ground Truth (GT)} \\
\hline
1.mp4 & ['W', 'AH', 'N', 'D', 'EY', 'AH', \textcolor{red}{'Y'}, 'NG', 'B', 'OY', 'K', 'AH', 'M', 'S', 'AH', 'P', 'AA', 'DH', 'AH', 'S', 'AH', 'N', 'F', 'L', 'AW', 'ER', 'W', 'AY', 'L', 'V', 'IH', 'S', 'IH', 'T', 'IH', 'NG', 'DH', 'AH', 'G', 'AA', 'R', 'D', 'AH', 'N', 'AH', 'N', 'D', 'HH', 'IY', 'N', 'OW', 'T', 'AH', 'S', 'AH', 'Z', 'HH', 'AW', 'W', 'IY', 'K', 'IH', 'T', 'L', 'UH', 'K', 'S'] & ONE DAY A YOUNG BOY COMES UP \textcolor{red}{ON} THE \textcolor{red}{SON FLOWER} WHILE VISITING THE GARDEN AND HE NOTICES HOW \textcolor{red}{WEEK} IT LOOKS & ONE DAY A YOUNG BOY COMES UPON THE SUNFLOWER WHILE VISITING THE GARDEN AND HE NOTICES HOW WEAK IT LOOKS \\ \hline
2.mp4 & ['JH', 'AH', 'S', 'T', 'L', 'AY', 'K', 'R', 'IY', 'CH', 'IH', 'NG', 'AW', 'T', 'T', 'UW', 'DH', 'AH', 'S', 'AH', 'N', 'F', 'L', 'AW', 'ER', 'M', 'AY', 'P', 'R', 'AH', 'V', 'AY', 'D', 'IH', 'NG', 'S', 'AH', 'M', 'W', 'AH', 'N', 'HH', 'UW', 'IH', 'Z', 'K', 'AH', 'G', 'L', 'EH', 'K', 'T', 'AH', 'D', 'AY', 'S', 'AH', 'L', 'EY', 'T', 'AH', 'D', 'AO', 'R', 'F', 'ER', 'G', 'AA', 'T', 'AH', 'N'] & JUST LIKE REACHING OUT TO THE \textcolor{red}{SON FLOWER} BY PROVIDING SOME \textcolor{red}{ONE} WHO IS NEGLECTED ISOLATED OR \textcolor{red}{FOR GOT TEN} & JUST LIKE REACHING OUT TO THE SUNFLOWER BY PROVIDING SOMEONE WHO IS NEGLECTED ISOLATED OR FORGOTTEN \\ \hline
\end{tabular}
\caption{Common errors in phoneme $\rightarrow$ word predictions, including homophones and compound words.}
\label{supptab1}
\end{table*}

\subsection{Part 2: Challenging Cases}
In Tab.~\ref{suptab2}, we highlight challenging cases such as the omission of {\ttfamily Kofi Annan}. This demonstrates the model's difficulty in reconstructing previously unseen names during phoneme-to-word mapping. Such issues could be mitigated with additional pretraining on larger and more diverse text datasets.

\begin{table*}[t]
\centering
\begin{tabular}{|c|p{5cm}|p{5cm}|p{5cm}|}
\hline
\textbf{File} & \textbf{Video $\rightarrow$ Phoneme (Errors in \textcolor{red}{red})} & \textbf{Phoneme $\rightarrow$ Word (Errors in \textcolor{red}{red})} & \textbf{Ground Truth (GT)} \\
\hline
6.mp4 & ['K', 'OW', 'F', 'IY', 'AE', 'N', 'AH', 'N', 'S', 'EH', 'D', 'DH', 'IH', 'S', 'W', 'IH', 'L', 'B', 'IY', 'M', 'EH', 'N', 'AH', 'F', 'IH', 'SH', 'AH', 'L', 'T', 'UW', 'M', 'AY', 'T', 'R', 'UW', 'P', 'Z', 'AA', 'N', 'DH', 'AH', 'G', 'R', 'N', 'D'] & \textcolor{red}{NAN} SAID THIS WILL BE BENEFICIAL \textcolor{red}{TOO} MY TROOPS ON THE GROUND & KOFI ANNAN SAID THIS WILL BE BENEFICIAL TO MY TROOPS ON THE GROUND \\ \hline
7.mp4 & ['AH', 'L', 'T', 'AH', 'M', 'AH', 'T', 'L', 'IY', 'DH', 'AE', 'T', 'S', 'W', 'AH', 'T', 'IH', 'T', 'Z', 'AH', 'AW', 'T'] & ULTIMATE \textcolor{red}{LEE} THATS WHAT ITS ABOUT & ULTIMATELY THAT'S WHAT IT'S ABOUT \\ \hline
\end{tabular}
\caption{Challenging cases, including omissions of names and complex phoneme $\rightarrow$ word mappings.}
\label{suptab2}
\end{table*}

\subsection{Part 3: Correct Predictions}
In Tab.~\ref{supptab3}, we showcase examples where the model successfully predicted the phoneme $\rightarrow$ word mappings without errors. These results demonstrate the model's capability to reconstruct accurate text from silent video inputs in scenarios with strong phoneme-word correlations and sufficient representation in the fine-tuning dataset.

\begin{table*}[t]
\centering
\begin{tabular}{|c|p{5cm}|p{5cm}|p{5cm}|}
\hline
\textbf{File} & \textbf{Video $\rightarrow$ Phoneme} & \textbf{Phoneme $\rightarrow$ Word} & \textbf{Ground Truth (GT)} \\
\hline
3.mp4 & ['IH', 'N', 'M', 'AY', 'F', 'EY', 'TH'] & IN MY FAITH & IN MY FAITH \\ \hline
4.mp4 & ['AY', 'TH', 'IH', 'K', 'DH', 'AH', 'K', 'AE', 'M', 'ER', 'AH', 'IH', 'S'] & I THINK THE CAMERA IS & I THINK THE CAMERA IS \\ \hline
5.mp4 & ['DH', 'IH', 'S', 'M', 'AH', 'S', 'T', 'B', 'IY', 'K', 'R', 'IY', 'EY', 'T', 'D'] & THIS MUST BE CREATED & THIS MUST BE CREATED \\ \hline
8.mp4 & ['AE', 'K', 'CH', 'UW', 'AH', 'L', 'IY', 'Y', 'UW', 'ER'] & ACTUALLY YOU ARE & ACTUALLY YOU ARE \\ \hline
9.mp4 & ['DH', 'EY', 'IH', 'N', 'V', 'EH', 'N', 'T', 'AH', 'D', 'DH', 'AE', 'T', 'T', 'R', 'AH', 'D', 'IH', 'SH', 'AH', 'N', 'F', 'AO', 'R', 'DH', 'EH', 'R', 'ER', 'AY', 'V', 'AH', 'L', 'HH', 'IY', 'R'] & THEY INVENTED THAT TRADITION FOR THEIR ARRIVAL HERE & THEY INVENTED THAT TRADITION FOR THEIR ARRIVAL HERE \\ \hline
10.mp4 & ['T', 'AY', 'D', 'IY', 'M', 'UW', 'T', 'S', 'IH', 'S', 'V', 'EH', 'R', 'IY', 'F', 'AH', 'S', 'IY', 'AH', 'B', 'AW', 'T', 'HH', 'IH', 'Z', 'F', 'UH', 'T', 'W', 'EH', 'R'] & TIDY BOOTS IS VERY FUSSY ABOUT HIS FOOTWEAR & TIDY BOOTS IS VERY FUSSY ABOUT HIS FOOTWEAR \\ \hline
\end{tabular}
\caption{Examples of correct phoneme $\rightarrow$ word predictions. These results showcase the model's ability to caption silent videos with high accuracy.}
\label{supptab3}
\end{table*}

\end{document}